\pdfoutput=1
\documentclass[11pt]{article}

\usepackage[table,xcdraw]{xcolor}
\usepackage{amsmath}
\usepackage{algorithm}
\usepackage{algpseudocode}
\usepackage{enumitem}

\usepackage{acl}

\usepackage{times}
\usepackage{latexsym}

\usepackage[T1]{fontenc}
\usepackage[utf8]{inputenc}

\usepackage{microtype}

\usepackage{hyperref}
\usepackage{url}
\usepackage{geometry}
\usepackage{booktabs}
\usepackage{tabularx}
\usepackage{graphicx} 
\usepackage{pifont}
\title{
Strategic Demonstration Selection for Improved \\ Fairness in LLM In-Context Learning}

\author{Jingyu Hu\\
  University of Bristol\\ Bristol, UK \\
  \text{ym21669@bristol.ac.uk} \\ \And
  Weiru Liu \\
  University of Bristol \\Bristol, UK \\
  \text{weiru.liu@bristol.ac.uk}  \\ \And
  Mengnan Du\\
  New Jersey Institute of Technology\\ Newark, USA \\
  \text{mengnan.du@njit.edu} \\}
\begin{document}
\maketitle
\begin{abstract}
Recent studies highlight the effectiveness of using in-context learning (ICL) to steer large language models (LLMs) in processing tabular data, a challenging task given the structured nature of such data. Despite advancements in performance, the fairness implications of these methods are less understood. This study investigates how varying demonstrations within ICL prompts influence the fairness outcomes of LLMs. Our findings reveal that deliberately including minority group samples in prompts significantly boosts fairness without sacrificing predictive accuracy. Further experiments demonstrate that the proportion of minority to majority samples in demonstrations affects the trade-off between fairness and prediction accuracy. Based on these insights, we introduce a mitigation technique that employs clustering and evolutionary strategies to curate a diverse and representative sample set from the training data. This approach aims to enhance both predictive performance and fairness in ICL applications. Experimental results validate that our proposed method dramatically improves fairness across various metrics, showing its efficacy in real-world scenarios.

\end{abstract}

\section{Introduction}

Large Language Models (LLMs), such as GPT-4~\citep{openaicite}, Claude-3~\citep{AnthropicAI2023}, and LLaMA-2~\citep{touvron2023llama2}, have achieved state-of-the-art performance in many natural language processing tasks.
These LLMs can adapt to different tasks by adding in-context prompts, without needing to retrain on the entire new dataset.
This optimization technique for LLMs is called in-context learning (ICL), which leverages specific input prompts to guide LLMs to generate more accurate outputs.
Recent research suggests that incorporating specially selected demonstrations into these prompts can significantly enhance LLM performance \citep{brown2020language, schick2020exploiting}.

Due to prompt length limitations, traditional LLMs have faced challenges in processing demonstrations from tabular datasets, which have a large number of features. However, with recent LLMs relaxing input length constraints, new avenues for applications in tabular datasets are opening up. \citep{hegselmann2023tabllm} has confirmed the predictive capabilities of LLMs on datasets from UCI repository.
Considering the usages of tabular data in high-stakes domains \citep{grinsztajn2022tree}, ensuring fairness alongside prediction performance is crucial for increasing trust in LLMs.
%Apart from prediction performance, ensuring fairness is also crucial for increasing trust in LLMs. 
\citep{liu2023investigating} has highlighted biases in LLM predictions with tabular datasets, but there are limited further investigations on how the fairness of LLMs performance varies with different ICL demonstrations.

To bridge this gap, we aim to answer the following research question: \emph{How do different demonstration strategies impact the fairness performance of LLMs on tabular classification tasks? And is there a demonstration strategy better than other strategies?}  
To better understand the impact of in-context learning on fairness, our proposed demonstration strategy considers the distribution of both demographic groups and target labels. A dataset can be divided into subgroups by demographic features, labelling the smallest as the minority (underrepresented) and the larger one as the majority. The fairness investigation compares differences between majority and minority groups. Our investigation includes evaluating five advanced LLMs, i.e., Text-davinci-003, GPT-3.5-turbo, GPT-4-turbo \footnote{\scriptsize\url{https://platform.openai.com/docs/models/}}, Claude-3 Haiku, and Claude-3 Sonnet\footnote{\scriptsize\url{https://www.anthropic.com/news/claude-3-family}}, across two fairness-focused tabular datasets: Credit and Adult. We found that prioritizing underrepresented samples and conscientiously including minority demographic groups and target labels during few-shot learning can significantly improve the fairness performance in LLMs output.

Despite the experimental observations, we are still wondering: \emph{Why \textcolor{black}{does} prioritizing minority group demonstrations benefit the fairness performance of LLMs in tabular-based classification tasks?} To further clarify this phenomenon, we perturb prediction labels and sensitive features in selected demonstrations and compare how the prediction outcomes of LLMs would be altered. Through these perturbation experiments, we found that increasing the proportion of underrepresented labels enhances fairness, but can lead to a decline in prediction performance, and vice versa.

Up until now, the above findings and explanations have been based on random demonstration selection. We hypothesize that: \emph{We can deliberately select demonstrations to further improve fairness performance.}
Motivated by the fiLter-thEn-Search (LENS) \citep{li2023finding} for textual classification, we adopt a similar process for extracting tabular demonstrations: first refine the training data set into a candidate pool and then sample and evaluate these candidates to identify the most supportive demonstrations. To this end, we introduced the 
\underline{F}airness via \underline{C}lustering-\underline{G}enetic (FCG) algorithm to effectively extract representative samples, to further enhance the fairness of LLM. Unlike LENS, which relies on progressive iterations on LLMs for candidate selection, our FCG method utilizes clustering. Clustering does not require access to LLMs and maintains diversity among the selected shots, effectively addressing concerns related to the time required for iterations and the cost of LLM access. 
Additionally, previous studies often assume the same selection probabilities for candidates across evaluation iterations, requiring enormous iterations to ensure that each sample is equally considered. Inspired by genetic evolution concepts, we adopt dynamic probabilities which give priority to representative samples with higher selection chances. Sample representativeness is measured by the LLM performance score, whose score is updated for each iteration. In this way, FCG can narrow the final sample set more efficiently, and drastically reduce the number of iterations needed.
We implement experiments to evaluate the proposed FCG demonstration selection method.
The results confirm that FCG algorithm improves LLMs performance in almost all strategies, with prioritizing the minority group still yielding the best results.

To conclude, the main contributions in this paper are as follows:
\begin{itemize}[leftmargin=*]\setlength\itemsep{-0.3em}
  \item We find that prioritizing underrepresented samples and conscientiously including minority demographic groups and target labels during few-shot learning can dramatically improve the fairness performance in LLM output (Section \ref{sec:diff_res_ss}). 

  \item We explain why prioritizing minorities leads to a fairer solution, and find the trade-off between LLMs' performance and demographic labels: increasing the ratio of underrepresented labels enhances fairness, but can lead to a decline in prediction performance, and vice versa (Section \ref{subsec:perturb}). 

  \item We propose the FCG (Fairness via Clustering-Genetic) algorithm, an efficient approach to retrieve a diverse and representative set of demonstrations from training data. Across almost all strategies, FCG enhances fairness in LLMs under in-context learning (Section \ref{sec:fairer-mitigation}).

\end{itemize}

\section{Experiment Setup}

Our primary goal is to investigate how different few-shot demonstration choices influence the fairness performance of LLMs under the in-context learning (ICL) setting. \textcolor{black}{Detailed related work on this area is discussed in Appendix \ref{app:related_work}}. In this section, we introduce the overall experimental setups.

\noindent
\textbf{Notations.}\, Given a dataset $D={(X,Y,Z)}_{i=1}^n$ where features $X \in \mathcal{R}$, the binary classification labels $Y \in \mathcal{Y} := \{0,1\}$, and sensitive feature $Z \in \mathcal{Z} := \{0,1\}$. We set $Z=0$ to represent the minority group and $Z=1$ as the majority group. $D$ is split into training dataset $D_{tr}$, validation dataset $D_{dev}$ and testing dataset $D_{test}$. 
\textcolor{black}{For each data point $d \in D := \{x, y, z \}$, a classifier $f$ predicts the label $f(x)$ based on the input features $x$.}

Given a subset $D' \in D_{tr}$, the proportion of samples where $Z=0$ within $D'$ is denoted as $r_z$. Specifically, $r_z=1$ means all samples in $D'$ belong to a minority group, whereas $r_z=0$ implies that every sample in $D'$ is from the majority group. Similarly, the proportion of samples for which $Y=0$ within $D'$ is represented by $r_y$.
%@todo CNT explain meaning.
\begin{equation}
\small
\label{alg:rzry}
r_z=\frac{\text{Count}(D'_{Z=0})}{\text{Count}(D')}, 
r_y=\frac{\text{Count}(D'_{Y=0})}{\text{Count}(D')} 
\end{equation}

\noindent
\textbf{Models and Datasets.}\, We use five LLMs \textcolor{black}{as $f$}: Text-davinci-003 (Davinci), GPT-3.5-turbo, GPT-4-turbo, Claude-3 Haiku, and Claude-3 Sonnet. The temperature in the model parameter is set to zero to ensure consistent responses. We select two tabular-based fairness datasets: default of credit card clients dataset (Credit, \citep{misc_default_of_credit_card_clients_350}) and adult income (Adult, \citep{misc_adult_2}). The Credit dataset covers information on credit card clients in Taiwan, including demographics, bills, payment history, etc. Its target is to predict whether there will be an overdue payment next month. The Adult dataset is to predict whether an individual's annual income exceeds 50K based on their individual features.
\textcolor{black}{Appendix \ref{app:dataset} contains further descriptions of dataset structures.}

\noindent
\textbf{Evaluation Metrics.}\, 
The predictive performance of LLMs on labels $Y$ is evaluated by metrics $Accuracy$, $Precision$, $Recall$, and \emph{F-score} \footnote{\url{https://scikit-learn.org/stable/}}.
We introduce $\Delta_{dp}$, $R_{dp}$, $\Delta_{eo}$, and $R_{eo}$ to evaluate fairness \footnote{\url{https://fairlearn.org/main/user_guide/fairness_in_machine_learning.html}}. They refer to the differences and ratios of Demographic Parity (DP) \cite{dwork2012fairness} and Equalized Odds (Eodds) \cite{hardt2016equality} between subgroups.

The demographic parity of the two groups partitioned by $Z$ is defined by Equation \ref{alg:DP0-DP1}. DP difference $\Delta_{dp}$ represents the difference between two, and DP ratio $R_{dp}$ is the ratio of the $DP_0$ and $DP_1$.
\begin{equation}
\small
\begin{aligned}
DP_0 &= P(f(x)=1 \mid Z=0) \\
DP_1 &= P(f(x)=1 \mid Z=1)
\end{aligned}
\label{alg:DP0-DP1}
\end{equation}

The True Positive Rate (TPR) and False Positive Rate (FPR) for both subgroups ($Z=0$ and $Z=1$) are defined as follows.

\begin{equation}
\small
\begin{aligned}
TPR_0 &= P(f(x) = 1 \mid y = 1, Z = 0) \\
TPR_1 &= P(f(x) = 1 \mid y = 1, Z = 1) \\
FPR_0 &= P(f(x) = 1 \mid y = 0, Z = 0) \\
FPR_1 &= P(f(x) = 1 \mid y = 0, Z = 1)
\end{aligned}
\end{equation}

\noindent
Eodds difference $\Delta_{eo}$ is defined as the greater metrics of TPR and FPR differences (Equation \ref{alg:eod}) where $\Delta TPR=(TPR_1-TPR_0)$ and $\Delta FPR =(FPR_1-FPR_0)$.
\begin{equation}
\small
\label{alg:eod}
\Delta_{eo}=\max(\Delta TPR, \Delta FPR)
\end{equation}
Eodds ratio $R_{eo}$ is the smaller ratio of TPR and the ratio of FPR between two groups, as shown below. 

\textcolor{black}{Here $\epsilon$ is used to avoid the setting where the denominator is zero, where we set $\epsilon = 1^{-6}$:}

\begin{equation}
\small
\label{alg:eor}
R_{eo}=\min \left(\frac{TPR_0}{TPR_1 + \epsilon}, \frac{FPR_0}{FPR_1 + \epsilon}\right)
\end{equation}

\textcolor{black}{The four fairness metrics range from 0 to 1. Lower $\Delta_{dp}$ and $\Delta_{eo}$ show smaller performance differences between groups, which points to fairer predictions. Higher $R_{eo}$ and $R_{dp}$ reflect more consistent performance across subgroups, suggesting better fairness.}

%\newpage

\begin{figure*}[t]
\begin{center}
\includegraphics[width=0.9\textwidth]{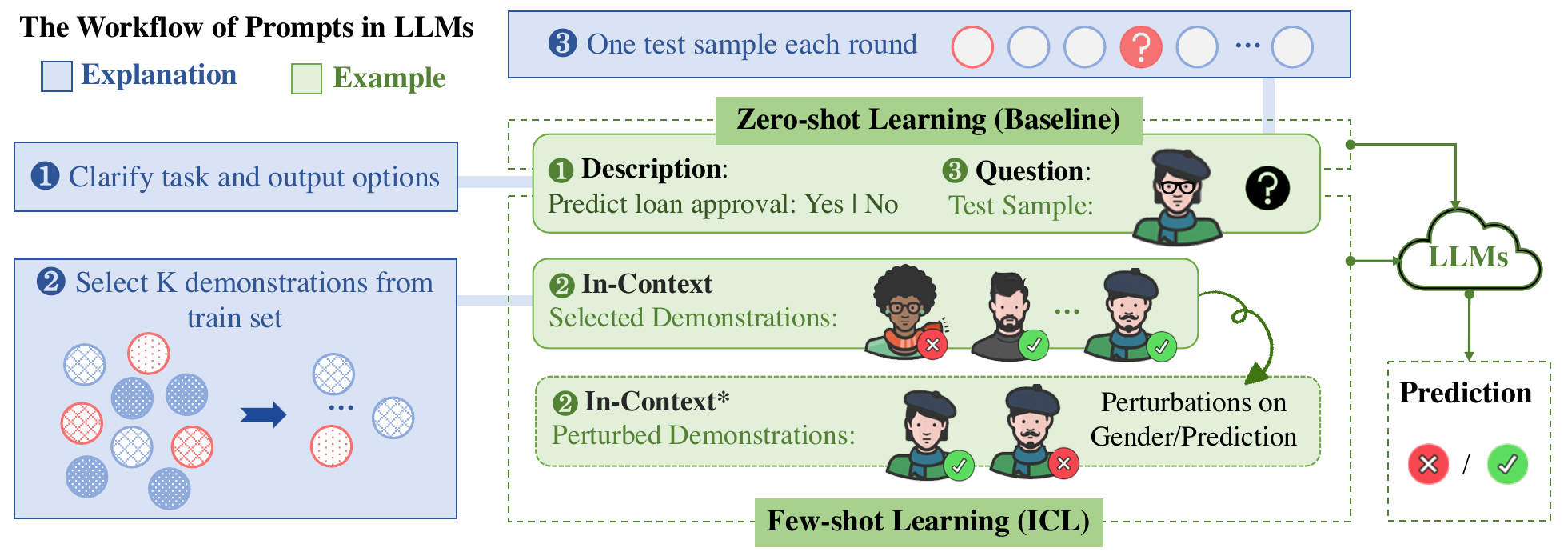}
\caption{The Prompt Template and Content Example (\textcolor{black}{*Perturbation is optional and is used to test the effectiveness of ICL. We discussed perturbations in Section \ref{subsec:perturb}})}
\label{fig:demo_prompt}
\end{center}
\end{figure*}

\noindent
\textbf{Prompt Template.} 
The output answer from the LLMs is based on the input prompt. As shown in Figure \ref{fig:demo_prompt}, the structure of a prompt can be divided into three parts: task description, in-context demonstrations, and questions. Part \ding{182} clarifies the task and defines the format of output prediction label options. Part \ding{183} contains a series of demonstrations as references. Part \ding{184} is the sample to be predicted.

We consider both zero-shot learning and few-shot learning in our experiments. Zero-shot learning refers to LLMs with a prompt exclude demonstration references (without \ding{183}) and is set as the baseline. Few-shot learning, sometimes also called in-context learning (ICL), consists of all three parts as input prompts. We compare how different demonstrations in part \ding{183} influence the fairness of LLMs. 
%Figure \ref{fig:demo_prompt} in green color gives a brief example of the prompt content. 
\textcolor{black}{The prompt example in Figure \ref{fig:demo_prompt} simplifies the tabular dataset, the detailed template is provided in Appendix \ref{app:prompt}.}

\section{How Demonstrations Impact Fairness of LLMs for Tabular Inputs?}
\label{sec:diff_res_ss}

\begin{figure*}[t]
\begin{center}
\includegraphics[width=1\textwidth]{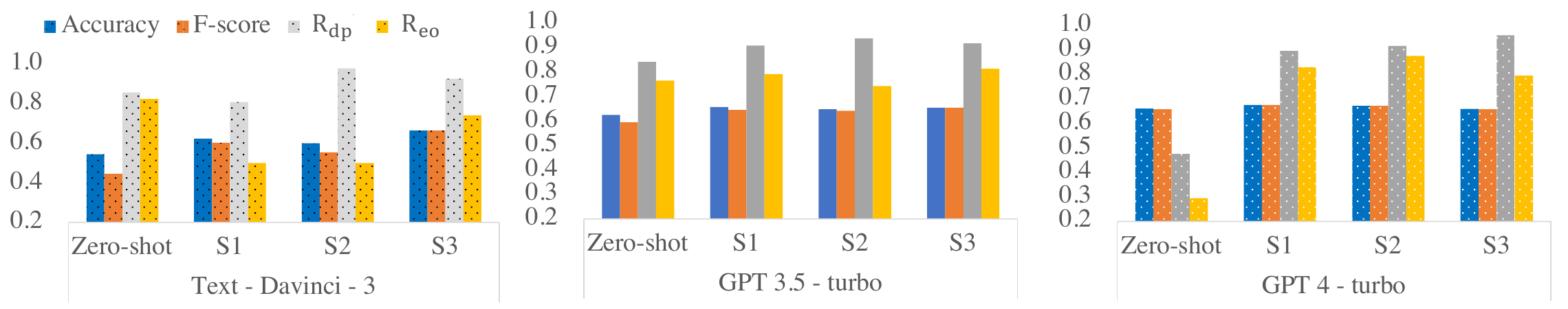}
\vspace{-15pt}
\caption{Prediction and fairness performance comparison across different LLMs on Credit dataset. It shows improvements in fairness metrics when samples from minority groups are prioritized. }
\label{fig:credit}
\end{center}
\end{figure*}

\begin{table*}[t]
\small
\begin{center}
\caption{Performance of GPT3.5-turbo with zero-shot and different few-shot strategies (S1, S2, S3) on Adult Income dataset. It demonstrates that strategic inclusion of demonstrations, particularly those from minority groups, can significantly enhance both predictive performance and fairness outcomes.}
\begin{tabular}{ccccc}
\toprule
Prediction & Zero-shot & $r_z=0.5, r_y=0.5$ (S1) & $r_z=1, r_y=0.5$ (S2)   & $r_z=1, r_y=1$ (S3)     \\
\midrule
\textbf{$Accuracy$} ↑  & 0.6855    & 0.7312 ± 0.0009 & \textbf{0.7328 ± 0.0028} & 0.7230 ± 0.0014 \\
\textbf{$Precision$} ↑ & \textbf{0.8519}    & 0.7936 ± 0.0012 & 0.7841 ± 0.0051 & 0.7808 ± 0.0038 \\
\textbf{$Recall$} ↑    & 0.4492    & 0.6250 ± 0.0012 & \textbf{0.6461 ± 0.0130} & 0.6122 ± 0.0036 \\
\textbf{$F$-$score$} ↑   & 0.5882    & 0.6993 ± 0.0011 & \textbf{0.7060 ± 0.0062} & 0.6915 ± 0.0017 \\
\toprule
Fairness   & Zero-shot & $r_z=0.5, r_y=0.5$ (S1) & $r_z=1, r_y=0.5$ (S2)   & $r_z=1, r_y=1$ (S3)     \\
\midrule
\textbf{$R_{dp}$} ↑   & 0.4063    & 0.6470 ± 0.0019 & 0.6769 ± 0.0080 & \textbf{0.6732 ± 0.0095} \\
\textbf{$R_{eo}$} ↑   & 0.1111    & 0.3682 ± 0.0044 & 0.4152 ± 0.0125 & \textbf{0.4722 ± 0.0187} \\ 
\textbf{$\Delta_{dp}$} ↓   & 0.2227    & 0.1688 ± 0.0009 & 0.1578 ± 0.0031 & \textbf{0.1555 ± 0.0046} \\
\textbf{$\Delta_{eo}$} ↓   & 0.3203    & 0.1875 ± 0.0019 & \textbf{0.1859 ± 0.0058} & 0.1906 ± 0.0071 \\
\bottomrule
\end{tabular}
\label{tab:diff-sd-adult}
\end{center}
\end{table*}

%\section{Fairness via Different Few-Shot Strategies}
In this section, we aim to answer: how do few shot demonstrations influence the fairness performance of LLMs for processing tabular inputs under the in-context learning setting?

To investigate this, we examine fairness performance variances across different demonstrations. We propose different combinations of prediction feature distribution $r_z$ and sensitive feature distribution $r_y$, expecting to explore the potential correlation between these feature distributions and LLM fairness. 
In the experiment, different demonstrations are based on three distinct sampling strategies denoted as $S1$, $S2$, and $S3$, each with unique distribution combinations of $r_z$ and $r_y$. 

\begin{itemize}[leftmargin=*]\setlength\itemsep{-0.3em}
    \item S1: Balanced Samples with Balanced Labels ($r_z=0.5, r_y=0.5$);
    \item S2: Prioritize Minority Samples with Balanced Labels ($r_z=1, r_y=0.5$);
    \item S3: Prioritize Minority Samples with Unbalanced Labels ($r_z=1, r_y \neq 0.5$).
\end{itemize}

Figure \ref{fig:credit} displays the performance of different LLMs on the Credit dataset. $r_y$ is set to 1 in S3. The fairness performance improves when prioritizing samples from minority groups ($r_z=1$) compared to a balanced sample selection ($r_z=0.5$).

Similar findings are found in the Adult dataset.
Table \ref{tab:diff-sd-adult} presents the performance of the GPT-3.5-turbo with zero-shot and different few-shot strategies. To ensure the stability and reliability of the results, we use random seeds set=\{25, 35, 42, 45, 55\} when selecting few-shot samples. The presented table summarizes average values and standard errors for the random seeds set. 

Overall, the results show that all few-shot strategies have generally improved fairness compared to zero-shot learning without lowering predictions. Also, prioritizing minorities (S2, S3) is an effective way to improve fairness. In contrast, balanced prompts (S1) show worse fairness performance.
\textcolor{black}{To further explain the observed pattern, we implement additional experiments and discussions on GPT-3.5's performance under the Adult dataset in the following sections. Complete results for other LLMs (e.g., Claude), using different seeds, are included in Appendix \ref{app:res}.}

\section{Why Prioritizing  Minority Group Demonstrations Benefit Fairness?}
\label{subsec:perturb}

The above analysis points out a strong correlation between prioritizing minority group demonstrations with improved fairness performance of LLMs. However, it is not yet clear how and why this phenomenon occurs.
\textcolor{black}{Thereby our next step is to clarify which part of the demonstrations most influenced the performance of LLMs.}
Specifically, we perturb the prediction label $Y$ and the sensitive feature $Z$ in selected demonstrations and compare how the prediction outcomes of LLMs would be altered.

\begin{figure}[t]
\begin{center}
\includegraphics[width=0.35\textwidth]{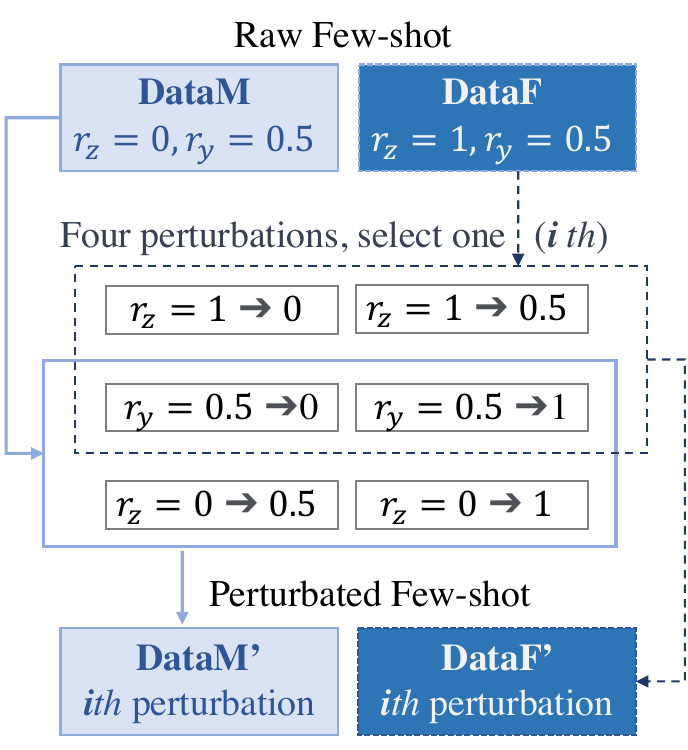}
\caption{The Workflow of Perturbations}
\label{fig:workflow_pertubation}
\end{center}
% \vspace{-10pt}
\end{figure}

The following experiment is performed on the Adult dataset with `income' as feature $Y$ and `gender' as feature $Z$. We set the random seed to 55 to extract two groups with balanced labels $DataF$ and $DataM$ from $D_{tr}$ as raw demonstrations. (1) $DataF$: balanced high-income and low-income females ($r_y=0.5, r_z=1$). (2) $DataM$: balanced high-income and low-income males ($r_y=0.5, r_z=0$). 
Figure \ref{fig:workflow_pertubation} illustrates the perturbation workflow. We define four perturbations, each consisting of the feature to be perturbed and the new proportions after perturbation. For example, perturbing $r_y=0.5 \rightarrow 1$ means that the quantity of high-income and low-income samples are balanced in raw demonstrations ($r_y=0.5$), and the perturbed demonstrations will become all low-income samples ($r’_y=1$) by flipping high-income labels to low-income.

The next part will discuss how perturbations at different proportions affect the overall prediction and fairness performance of LLM, along with a deeper performance comparison within subgroups.

\subsection{Perturbations Impact on Overall Fairness}

Table \ref{tab:data_pertu} compares the prediction and fairness performance with different perturbations on gender and income.

%In income perturbations, 
Perturbing the income labels for $DataF$ and $DataM$ leads to a certain degree of decline in predictive performance ($1\%$ to $6\%$). 
\citealp{min2022rethinking} mentioned a similar phenomenon that replacing gold labels with random labels only presents the marginal impact on performance. 
Nevertheless, we also found that altering the ground truth labels (income) can greatly affect fairness performance, resulting in a drastic drop in all scenarios.
Especially when replacing labels with high income, the $R_{dp}$ in DataF decreased from 71.32\% to 50.54\%, and in DataM, it decreased from 50.94\% to 43.06\%.

When we perturbed gender labels, results show that fairness performance improves with a higher proportion of females. The fairness performance decreases when we perturb from female to male in $DataF$, as $R_{dp}$ decreases from 71.32\% to 59.15\%. Similar patterns are observed in $DataM$, where fairness gradually increases by 8.1\% when modifying from male to female.

\textcolor{black}{In most cases, the perturbation results align with the intuition that distorting real data can degrade its quality, thus potentially leading to negative impacts on LLMs performance. However, we also find that perturbing to a higher ratio of minority labels ($r_z'=1$) can positively enhance fairness, suggesting a strong connection between fairness performance and sensitive labels. To further validate this finding, Section \ref{sec:per_subgroups} compares performance variations at the subgroup level.}
\begin{table*}[t]
\begin{center}
\caption{Prediction and Fairness Performance on Income and Gender Perturbations}
\label{tab:data_pertu}
\resizebox{\textwidth}{!}{
\begin{tabular}{@{}cccc|ccc|ccc|cccc@{}}
\toprule
& \multicolumn{6}{c|}{Different perturbations on income} & \multicolumn{6}{c}{Different perturbations on gender} \\
\rowcolor[HTML]{EFEFEF} 
           & \multicolumn{3}{c|}{$r_z=1, r_y=0.5$ (DataF)}           & \multicolumn{3}{c|}{$r_z=0, r_y=0.5$ (DataM)} & \multicolumn{3}{c|}{$r_z=1, r_y=0.5$ (DataF)} & \multicolumn{3}{c}{$r_z=0, r_y=0.5$ (DataM)}          \\ 
\midrule
\rowcolor[HTML]{EFEFEF} 
Prediction & Raw &   $ r'_y=1$ &  $r'_y=0$ & Raw &  $ r'_y=1$ & $r'_y=0$ & Raw & $r'_z=0.5$ & $r'_z=0$ & Raw & $r'_z=0.5$ & $r'_z=1$\\
$Accuracy$ ↑  & \textbf{0.7480} & 0.7383 & 0.6992 & \textbf{0.7148} & 0.6914 & 0.6543 & 0.7480 & 0.7422 & \textbf{0.7617} & 0.7148 & \textbf{0.7168} & 0.7090\\
$Precision$ ↑ & 0.7873 & 0.7933 & \textbf{0.8643} & 0.8438 & 0.8451 & \textbf{0.8835} & 0.7873 & 0.7925 & \textbf{0.7965} & \textbf{0.8438} & 0.8364 & 0.8204\\
$Recall$ ↑    & \textbf{0.6797} & 0.6445 & 0.4727 & \textbf{0.5273} & 0.4688 & 0.3555 & 0.6797 & 0.6563 & \textbf{0.7031} & 0.5273 & 0.5351 & \textbf{0.5352}\\
$F$- $score$ ↑    & \textbf{0.7296} & 0.7112 & 0.6111 & \textbf{0.6490} & 0.6030 & 0.5070 & 0.7296 & 0.7179 & \textbf{0.7469} & 0.6490 & \textbf{0.6556} & 0.6478\\
\rowcolor[HTML]{EFEFEF} 
Fairness & Raw &   $ r'_y=1$ &  $r'_y=0$ & Raw &  $ r'_y=1$ & $r'_y=0$ & Raw & $r'_z=0.5$ & $r'_z=0$ & Raw & $r'_z=0.5$ & $r'_z=1$\\
$R_{dp}$ ↑      & \textbf{0.7132} & 0.6508 & 0.5054 & \textbf{0.5094} & 0.4639 & 0.4306 & \textbf{0.7132} & 0.6308 & 0.5915 & 0.5094 & 0.5421 & \textbf{0.5905}\\
$R_{eo}$ ↑      & \textbf{0.3824} & 0.3438 & 0.0556 & 0.1364 & \textbf{0.1579} & 0.0000 & \textbf{0.3824} & 0.2571 & 0.1500 & 0.1364 & 0.2273 & \textbf{0.3043} \\ 
$\Delta_{dp}$ ↓      & \textbf{0.1445} & 0.1719 & 0.1797 & 0.2031 & 0.2031 & \textbf{0.1602} & \textbf{0.1445} & 0.1875 & 0.2266 & 0.2031 & 0.1914 & \textbf{0.1680}\\
$\Delta_{eo}$ ↓      & \textbf{0.1641} & 0.1797 & 0.226 & 0.2578 & 0.2813 & \textbf{0.2266} & \textbf{0.1641} & 0.2031 & 0.2656 & 0.2578 & 0.2500 & \textbf{0.2109} \\
\bottomrule
\end{tabular}
}
\end{center}
\end{table*}

\subsection{Perturbations Impact across Subgroups}
\label{sec:per_subgroups}
Table \ref{tab:combined_pertu_ch} displays the model performance of TPR and FPR on both minority (female) and majority (male) subgroups after income and gender perturbations. \textcolor{black}{Similar to $DP$ and $EO$, the metrics $\Delta TPR$ and $\Delta FPR$ assess performance disparities between female and male subgroups. 
Equal treatment is achieved when these differences approach zero, hence, lower values of $\Delta TPR$ and $\Delta FPR$ are preferable. We also consider absolute values of FPR and TPR within each subgroup to fully assess fairness changes in perturbations.}

In income perturbations, replacing the income labels results in a decrease in TPR and FPR for both female and male groups, with a more significant decline observed in the female group. This reduction is most notable when income labels are changed to high-income. \textcolor{black}{In a few cases, the relative metrics $\Delta TPR$ and $\Delta FPR$ show improvement compared to the results without perturbations. However, the corresponding absolute metrics TPR and FPR do not show consistent trends and worsen instead. This inconsistency makes it difficult to validate the impact of income on fairness performance. Therefore, we conclude that the ground truth labels in the demonstrations are not the source of benefit for LLMs' fairness.}

In gender perturbations, however, \textcolor{black}{subgroup performance is greatly affected by these gender label changes. For absolute values,} flipping female labels to male in $DataF$ leads to a 4.69\% increase in FPR and a 5.47\% increase in TPR for the male group. Similarly, transforming male labels to female in $DataM$ results in increases in both TPR and FPR for the female group. \textcolor{black}{Similar trends are found in their relative values. Increasing the proportion of male labels leads to higher $\Delta TPR$ and $\Delta FPR$, illustrating greater difference in subgroup treatment. Conversely, an increase in the ratio of female labels leads to reductions in both $\Delta TPR$ and $\Delta FPR$, suggesting enhanced fairness.}

In general, the above results show a trade-off between the LLMs performance and demographic labels. LLMs exhibit improved performance when the proportion of minority groups increases: they become fairer compared \textcolor{black}{to the used of original data} when perturbing demographic labels male to female. Therefore, we conclude that prioritizing demonstrations from minority groups can maximize these advantages and promote fairness in LLMs. In contrast, perturbing labels leads to demonstrations becoming less reliable, as they can lead models to learn noise and perform worse. The perturbation on prediction labels (income) conforms with this pattern.

\begin{table*}[t]
\caption{TPR and FPR Assessment across Subgroups on Income and Gender Perturbations}
\centering
\resizebox{\textwidth}{!}{% Adjusting the size to fit within the text width
\begin{tabular}{@{}cccc|ccc|ccc|cccc@{}} % Added the vertical line here
\toprule
& \multicolumn{6}{c|}{Different perturbations on income} & \multicolumn{6}{c}{Different perturbations on gender} \\
\midrule
% & \multicolumn{6}{c|}{$r_z=1, r_y=0.5$ (DataF)}                    & \multicolumn{6}{c}{$r_z=0, r_y=0.5$ (DataM)}                    \\ \midrule
           & \multicolumn{3}{c|}{$r_z=1, r_y=0.5$ (DataF)}           & \multicolumn{3}{c|}{$r_z=0, r_y=0.5$ (DataM)} & \multicolumn{3}{c|}{$r_z=1, r_y=0.5$ (DataF)} & \multicolumn{3}{c}{$r_z=0, r_y=0.5$ (DataM)}          \\ 
\midrule
\rowcolor[HTML]{EFEFEF} 
 & Raw & $r'_y=1$ & $r'_y=0$ & Raw & $r'_y=1$ & $r'_y=0$ & Raw          & $r'_z=0.5$ & $r'_z=0$ & Raw          & $r'_z=0.5$ & $r'_z=1$ \\
 %\midrule
$TPR_{M}$ & \textbf{0.7422}  & 0.7344  & 0.5859  & \textbf{0.6563}  & 0.6094  & 0.4688  & 0.7422           & 0.7422            & \textbf{0.7969}           & \textbf{0.6563}           & 0.6641            & 0.6406          \\
$TPR_{F}$ & \textbf{0.617}2  & 0.5547  & 0.3594  & \textbf{0.3984}  & 0.3281  & 0.2422  & \textbf{0.6172}           & 0.5703            & 0.6094           & 0.3984           & 0.4141            & \textbf{0.4297}          \\
%\midrule
\rowcolor[HTML]{EFEFEF} 
$\Delta_{TPR}$ ↓ & \textbf{0.1250} & 0.1797 & 0.2266 & 0.2578 & 0.2813 & \textbf{0.2266} & \textbf{0.1250} & 0.1719  & 0.1875 & 0.2578 & 0.2500  & \textbf{0.2109} \\
%\midrule
$FPR_{M}$  & \textbf{0.2656}  & 0.2500  & 0.1406  & \textbf{0.1719}  & 0.1484  & 0.0938  & 0.2656           & 0.2734            & \textbf{0.3125}           & \textbf{0.1719}           & \textbf{0.1719}            & \textbf{0.1797}          \\
$FPR_{F}$ & \textbf{0.1016}  & 0.0859  & 0.0078  & \textbf{0.0234}  & \textbf{0.0234}  & 0.0000  & \textbf{0.1016}           & 0.0703            & 0.0469           & 0.0234           & 0.0391            & \textbf{0.0547}          \\
%\midrule
\rowcolor[HTML]{EFEFEF} 
$\Delta_{FPR}$ ↓ & 0.1641 & 0.1641 & \textbf{0.1328} & 0.1484 & 0.1250 & \textbf{0.0938} & \textbf{0.1641} & 0.2031  & 0.2656 & 0.1484 & 0.1328  & \textbf{0.1250} \\ 
\bottomrule
\end{tabular}
}
\label{tab:combined_pertu_ch}
\end{table*}

\section{Mitigation Algorithm for Fair Demonstration Selection}
\label{sec:fairer-mitigation}

The above results confirm that the application of diverse demonstrations, particularly those from the minority, can drastically influence the fairness of LLMs. 
Experiments on different sets of selected shots under the same strategy also reveal a similar trend, albeit with different absolute performance values. This leads to the question: \emph{how to extract representative demonstrations that yield better performance among these sets?}

Enumerating and evaluating the outcomes of LLMs across all possible sets is impractical due to the sheer number of combinations. 
Thus, in this section, we propose a fairness via clustering-genetic (FCG) algorithm to efficiently select influential demonstrations, leading LLMs to a better performance without having to explore all possible combinations.
The core idea of FCG includes three aspects: (1) \textcolor{black}{Use clustering to shrink the selection sets while maintaining diverse samples.} (2) Define a score that considers both prediction and fairness performance, applying a genetic approach to iteratively select and score these samples within the sets. (3) Rank samples from highest to lowest based on their scores to select the most influential ones.

\subsection{The Proposed FCG Algorihtm}

We introduce details of the proposed FCG mitigation algorithm in this section.

\begin{algorithm}[t]
\small
\caption{FCG Algorithm}
\begin{algorithmic}[1]

\Procedure{Step1: Diverse Clustering}{}
    \State all\_idx\_set =[]
    \For{each $g_i$ in $SG$}
        \State selected\_idx\_set = []
        \State X = $g_i$.get\_all\_data()
        \State centroids = Kmeans(clusters = n).fit($X$)
        \For{each center in centroids}
            %\State dis\_set = $||$ X - center $||_2$
            \State dis\_set = distance(X,center)
            \State closest = argsort(dis\_set)[:m]
            \State selected\_idx\_set.extend(closest)
        \EndFor
        \State all\_idx\_set.append(selected\_idx\_set)
    \EndFor
    \State \Return SG'.set(all\_idx\_set)
%\EndFunction
\EndProcedure

\Procedure{Step2:Update evol score}{}
\State ch = $[M_{pred}$, $M_{fair}$]
    \For{each $g_i'$ in SG'}
            \State shots = $g_i'$.genetic\_algorithm(k)  
            \State evol\_score = CAL\_SCORE(shots,ch)
            \State shots.update(evol\_score)
            \State repeat $iters$ times
        %\EndFor
    \EndFor
\EndProcedure

\Function{cal\_score}{shots,ch}
    \State $Y_{base}$= LLM.predict($X_{dev}$)
    \State $Base_{pred}, Base_{fair}$ = eval($Y_{base}$,$Y_{dev}$, ch)

    \State $Y_{ICL}$ = LLM.predict(shots, $X_{dev}$)
    \State $ICL_{pred}, ICL_{fair}$ = eval($Y_{ICL}, Y_{dev},$ ch)

    \State $\Delta pred = \max((ICL_{pred} - Base_{pred}), p)$
    \State $\Delta fair = \max((ICL_{fair} - Base_{fair}), p)$
    \State \Return $\alpha \cdot \Delta pred + (1-\alpha) \cdot \Delta fair$
\EndFunction
\end{algorithmic}
\label{alg:FCG}
\end{algorithm}

\noindent\textbf{Clustering.}\,
Based on the value combinations of the sensitive feature Z and label Y, we divide the training data $D_{tr}$ into four subgroups denoted as SG$=\{g_1(Z=1,Y=0), g_2(Z=1,Y=1),$$ g_3(Z=0,Y=0), g_4(Z=0,Y=1)\}$. 
For each subgroup, we apply k-means clustering to extract a diverse and representative initial population.  Each subgroup in $SG$ is clustered into $n$ clusters, with $m$ neighbors selected around each centre of the cluster. 
The filtered new subgroups are denoted as SG'=$\{g_1',g_2',g_3',g_4'\}$.

\noindent\textbf{Genetic Evolution for Score Updates.}\,
Next, for each subgroup within SG', we select K-demonstrations for $iters$ times using the roulette wheel genetic evolutionary approach and validate their ICL performance on $D_{dev}$.
The evolutionary method means that data with a higher score is more likely to be chosen in each round. The score is first set as the default initialisation score $p$ and then updated as the average of EvolScore computed during the iterations when the sample is selected. EvolScores integrates the performance of both prediction $M_{pred}$ and fairness $M_{fair}$ metrics, with $\alpha$ serving as the trade-off coefficient. The metrics provide options for selecting either $Accuracy$ or $F$-$score$ as $M_{pred}$, and either $R_{dp}$ or $R_{eo}$ as $M_{fair}$. EvolScores in $SG'$ will be updated and then used for subsequent selecting iterations. 

%Algorithm \ref{alg:FCG} provides the details of the whole process of FCG algorithm.

\begin{figure*}[t]
\begin{center}
\includegraphics[width=1\textwidth]{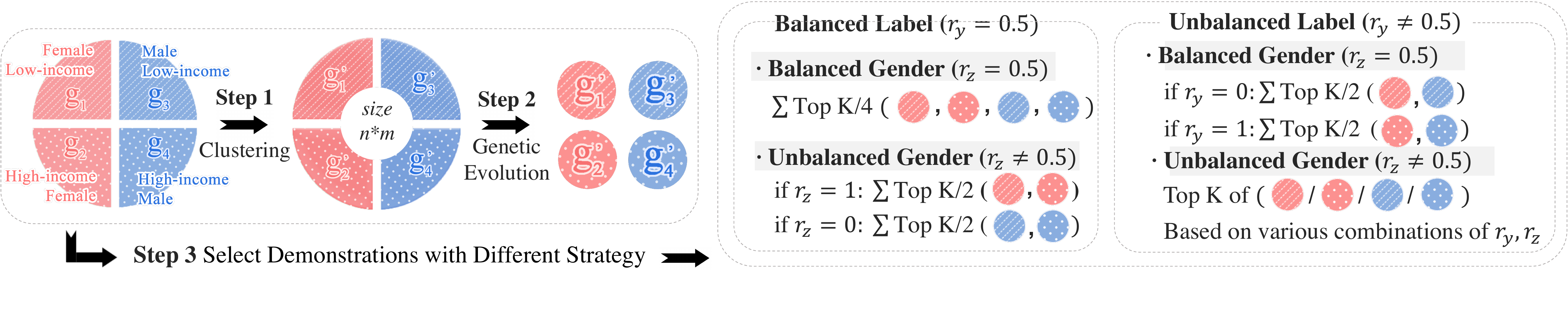}
\caption{The Workflow of Fairness via Clustering-Genetic (FCG) on the Adult Dataset (\textcolor{black}{$r_y=1$, all high-income; $r_y=0$, all low-income; $r_y=0.5$, balanced samples of high-income and low-income; $r_z=1$, all females; $r_z=0$, all males; $r_z=0.5$, balanced samples of females and males.})}
\label{fig:CGA_Example}
\end{center}
\end{figure*}

In the testing stage on $D_{test}$, demonstrations in $SG'$ are ranked by their average EvolScores, enabling different ICL strategies to select the top-performing demonstrations from their corresponding subgroups.
The detailed steps of FCG pseudocode is given in Algorithm \ref{alg:FCG}.
Figure~\ref{fig:CGA_Example} gives an example of the whole process of representative sample selections with FCG on Adult dataset.

\subsection{Experimental Results}
Table~\ref{tab:res_cga} presents the experimental results evaluating the debiasing performance of the proposed FCG algorithm.
The experiments are performed on the Adult dataset, setting the number of clusters to $n=8$ and the number of neighbors to $m=5$. We start with an initial score of $p=0.05$ and perform 10 iterations to update EvolScore. $F$-$score$ and $R_{eo}$ are selected for calculating Evolscore and the balancing parameter $\alpha$ is set to 0.5.

\begin{table*}[t]
\caption{The comparative analysis of the predictive and fairness performance achieved by various LLMs with demonstrations selected using the proposed FCG algorithm. The experiments are conducted on the Adult dataset. The table highlights that the FCG algorithm enhances fairness across almost all strategies.}
\begin{center}
\resizebox{0.98\textwidth}{!}{
\begin{tabular}{cccccccccccc}
\toprule
          &     Zero-shot     & \multicolumn{5}{c}{K-Shot (K=8)}                  & \multicolumn{5}{c}{K-Shot (K=4)}                  \\
\midrule
 & Zero-shot  & \multicolumn{3}{c}{$r_y=0.5$ (Balanced Labels)} & $r_y=1$ & $r_y=0$ & \multicolumn{3}{c}{$r_y=0.5$ (Balanced Labels)} & $r_y=1$ & $r_y=0$ \\
 \rowcolor[HTML]{EFEFEF}
Prediction  & Baseline & $r_z=0.5$ & $r_z=0$  & $r_z=1$  & $r_z=1$  & $r_z=1$  & $r_z=0.5$ & $r_z=0$  & $r_z=1$  & $r_z=1$  & $r_z=1$  \\
$Accuracy$ ↑  & 0.6855   & 0.7344  & 0.7363 & \textbf{0.7793} & 0.7500 & 0.7754 & 0.7656  & 0.7500 & 0.7656 & 0.7539 & 0.7578 \\
$Precision$ ↑ & \textbf{0.8519}   & 0.7174  & 0.6997 & 0.7360 & 0.7207 & 0.7640 & 0.7297  & 0.7222 & 0.7194 & 0.7338 & 0.7200 \\
$Recall$  ↑  & 0.4492   & 0.7734  & 0.8281 & \textbf{0.8711} & 0.8164 & 0.7969 & 0.8438  & 0.8125 & \textbf{0.8711} & 0.7969 & 0.8438 \\
$F$-$score$ ↑ & 0.5882   & 0.7444  & 0.7585 & \textbf{0.7979} & 0.7656 & 0.7801 & 0.7826  & 0.7647 & 0.7880 & 0.7640 & 0.7770 \\
\rowcolor[HTML]{EFEFEF}
Fairness  & Baseline & $r_z=0.5$ & $r_z=0$  & $r_z=1$  & $r_z=1$  & $r_z=1$  & $r_z=0.5$ & $r_z=0$  & $r_z=1$  & $r_z=1$  & $r_z=1$  \\
$R_{dp}$ ↑   & 0.4063   & 0.7692  & 0.7719 & \textbf{0.8938} & 0.7576 & 0.7919 & 0.7515  & 0.8000 & 0.8675 & 0.7707 & 0.8072 \\
$R_{eo}$ ↑   & 0.1111   & 0.6250  & 0.5690 & 0.7021 & 0.5577 & 0.5750 & 0.5094  & 0.6000 & \textbf{0.7059} & 0.5417 & 0.6800 \\ 
$\Delta_{dp}$ ↓   & 0.2227   & 0.1406  & 0.1523 & \textbf{0.0664} & 0.1563 & 0.1211 & 0.1641  & 0.1250 & 0.0859 & 0.1406 & 0.1250 \\
$\Delta_{eo}$ ↓   & 0.3203   & 0.1406  & 0.1953 & \textbf{0.1094} & 0.1797 & 0.1328 & 0.2031  & 0.1563 & 0.1172 & 0.1719 & 0.1250 \\
\bottomrule
\end{tabular}
}
\label{tab:res_cga}
\end{center}
\end{table*}

%-changed-
\textcolor{black}{Results show that demonstrations selected by FCG perform well, and greatly outperforming random sampling.}
It is worth noting that using a balanced set from minority samples continues to yield the best performance, proving our finding that prioritizing minority samples ($r_z=1$) remains an effective strategy in ICL.
Besides the minority group, the improvements in accuracy and fairness also happen in the majority group, which affirms the value of considering both factors in FCG selections.

\noindent
\textbf{Ablation Study.}\,
We implement ablation experiments to verify the utility of the two-step extracting process in FCG mitigation. In the ablation study, part of the samples are selected using the same flow of choosing the top K samples based on their EvolScores, while the other part is selected randomly. The results in Table \ref{tab:ablation} suggest: (1) Even when EvolScores are ignored when selecting partial samples, the results still outperform the raw random selection method (Random ($g_1+g_2$)), thus proving the effectiveness of the clustering selection in the first stage. (2) Moreover, both ablation test FCG ($g_1$) and FCG ($g_2$) performed worse compared to the results using complete FCG ($g_1+g_2$), further confirming the need for the second stage of genetic selection based on EvolScore scoring.

\begin{table}[t]
\caption{The Ablation Study of FCG under Balanced Labels in Minority Group Strategy (S2) for Selecting K Demonstrations (K=8). \textcolor{black}{S2 strategy is based on minority group ($z=1$) with two possible labels $y=\{0,1\}$, The corresponding subgroups are denoted as $g_1(z=1,y=0)$ and $g_2(z=1,y=1)$.}}
\resizebox{0.48\textwidth}{!}{
\begin{tabular}{ccccc}
\toprule
Shots from $g_1$ & Random & Top K/2 & Top K/2 & Random K/2 \\
\midrule
Shots from $g_2$ & Random & Top K/2 & Random K/2 & Top K/2 \\
\midrule
\rowcolor[HTML]{EFEFEF}
Prediction & Random ($g_1+g_2$) & FCG ($g_1+g_2$) & FCG ($g_1$) & FCG ($g_2$) \\
$Accuracy$ ↑ & 0.7480 & \textbf{0.7793} & 0.7500 & 0.7559 \\
$Precision$ ↑ & \textbf{0.7592} & 0.7360 & 0.7013 & 0.7079 \\
$Recall$ ↑ & 0.7266 & \textbf{0.8711} & \textbf{0.8711} & \textbf{0.8711} \\
$F score$ ↑ & 0.7425 & \textbf{0.7979} & 0.7770 & 0.7811 \\
\rowcolor[HTML]{EFEFEF}
Fairness & Random ($g_1+g_2$) & FCG ($g_1+g_2$) & FCG ($g_1$) & FCG ($g_2$) \\
$R_{dp}$ ↑ & 0.7254 & \textbf{0.8938} & 0.8276 & 0.8208 \\
$R_{eo}$ ↑ & 0.4390 & \textbf{0.7021} & 0.6964 & 0.6140 \\
$\Delta_{dp}$ ↓ & 0.1523 & \textbf{0.0664} & 0.1172 & 0.1211 \\
$\Delta_{eo}$ ↓ & 0.1797 & \textbf{0.1094} & 0.1328 & 0.1719 \\
\bottomrule
\end{tabular}
}
\label{tab:ablation}
\end{table}

\section{Conclusions}

In this paper, we investigate how the choice of demonstrations impacts the fairness of LLMs on tabular data classification tasks when using in-context learning. Through experiments, we found that prioritizing underrepresented groups and including minority examples in the few-shot demonstrations can significantly enhance fairness performance, without sacrificing predictive accuracy. Further analysis revealed that increasing the proportion of underrepresented labels improves fairness metrics like demographic parity and equal odds. To efficiently retrieve effective demonstrations, we proposed the FCG algorithm that uses clustering and genetic evolution to select a diverse and representative set of examples from the training data. Across multiple strategies and datasets, experimental results indicate that FCG was able to improve fairness compared to random sampling.

\clearpage
\section{Limitations}

While our study presents significant advancements in understanding and improving fairness in LLMs using in-context learning (ICL), several limitations should be noted. Firstly, we equally weigh fairness and prediction performance in evaluating representative demonstrations using our Fairness via Clustering-Genetic (FCG) algorithm, which might not align with real-world applications that require a dynamic balance between these metrics. Additionally, our focus on binary classification with a single sensitive feature limits the broader applicability of our findings. In future, we plan to explore LLM's intersectional fairness and its performance in multi-classification tasks. Lastly, while we used pre-trained models without fine-tuning, investigating how fine-tuning on curated samples impacts fairness could provide deeper insights.

\bibliography{anthology,custom}
\bibliographystyle{acl_natbib}

\clearpage
\appendix

\section{Related Work}
\label{app:related_work}
\subsection{Fairness in Large Language Model}
Addressing social biases is crucial for ensuring the trustworthiness of language models~\cite{nangia2020crowspairs, nadeem2020stereoset}. LLMs face similar fairness issues: many studies have confirmed that LLMs can capture social biases from unprocessed training data and transmit these biases to downstream tasks \cite{abid2021persistent, brown2020language, wang2023decodingtrust}.
\citet{abid2021large} addressed the issue of GPT-3's output displaying bias to Muslims. 
\citet{huang2021uncovering} found that bias in LLMs' responses exists even without prompts explicitly asking about it.
\citet{liang2023holistic} tested bias and stereotypes on LLMs using the BBQ dataset for question answering, finding that most models exhibit biases distinct from broader societal trends. \citet{wang2023decodingtrust} assesses bias by prompting GPT models to express their views on stereotypes. 
%While the majority of authors focus on biases in generated texts from LLMs
Most bias studies have focused on representational harms caused by LLM generations, with only limited studies \cite{hegselmann2023tabllm} addressing fairness concerns classification problems with tabular datasets.
%This paper examines effect of shots on the fairness of LLMs under NLP classification tasks
Besides investigation on pre-trained LLMs, recent research also focuses on ensuring fairness in other trained machine learning models, such as perturbation-based \cite{zhang2022fairness,wang2022fairness} and boosting-based \cite{kim2019multiaccuracy} approaches. %approaches are proposed to calibrate ML models to achieve fairness without retraining.

\subsection{In Context Learning}
In-context learning (ICL), known as few-shot learning, allows LLMs to learn tasks with minimal examples as demonstrations \cite{dong2022survey, zhao2021calibrate}. Positive impacts of ICL on LLMs have been observed in different tasks such as text classification and answering \cite{gao2021making,liu2021makes}, images generations \cite{bar2022visual}, speech tasks \cite{zhang2023speak}, and multi-modal scenarios \cite{huang2023language, wei2022emergent}.

Meanwhile, researchers have found that the performance of ICL is highly sensitive to the demonstration prompt \cite{chen2023relation, lu2021fantastically, zhao2021calibrate,shi2022xricl}. Investigations have explored factors that can influence ICL prediction performance, including demonstration retrievals \cite{tanwar2023multilingual, sorensen-etal-2022-information}, orderings \cite{lu-etal-2022-fantastically}, and input-label mapping \cite{yoo2022groundtruth, workrethinking}. 
%Most studies consider prediction performance when comparing different ICL strategies, but fairness is often overlooked when selecting 

\noindent
\textbf{Demonstration Retrievals.}\,
A common demonstration retrievals approach in ICL  involves randomly selecting a subset of examples from the training set \cite{brown2020language}. Given the sensitivity of LLMs to the prompts, there has been investigation into selecting representative samples to enhance outcomes. 
Selecting the top-K training examples is one mitigation method and has been demonstrated in semantic parsing \cite{rubin2021learning} and semantic classification \cite{chang2023data}. LENS \cite{li2023finding} proposed a two-step filter and search algorithm to identify informative support examples. Despite these advances, these retrieval techniques often focus solely on prediction performance and overlook the aspect of fairness. Additionally, most retrieval methods often require extensive experimental iterations, with significant time and resource investment.

\section{Dataset}
\label{app:dataset}

Table \ref{tab:app-default-dataset} describes the data structure in the Default Credit dataset. We calculate the mean values of PAY\_AMT\_i and BILL\_AMT\_i, and merge them into Avg\_PAY\_AMT and Avg\_BILL\_AMT separately. 
The raw Adult dataset shown in Table \ref{tab:app-adult-dataset} contains 14 features, excluding education-num, fnlwgt, race, and native-country for this experiment. `$>50K$' and `$\leq50K$' is mapped to `greater than 50K' and `less than or equal to 50K' respectively, for better alignment with the language model. In analysis, we call high income if the person's annual income is over 50K and low income if it is less than 50K.
The size ratio of $D_{train}$: $D_{dev}$: $D_{test}$ is 9:1:10 in both two datasets.
$K$ demonstrations are extracted from $D_{train}$, 60 samples are extracted from $D_{dev}$, 512 samples for $D_{test}$.
We consider the balanced group and balanced labels scenario and extract samples with parameter random\_seed=42.

\begin{table*}[t]
\begin{center}
\caption{Default Credit Dataset Description}
\includegraphics[width=0.8\textwidth]{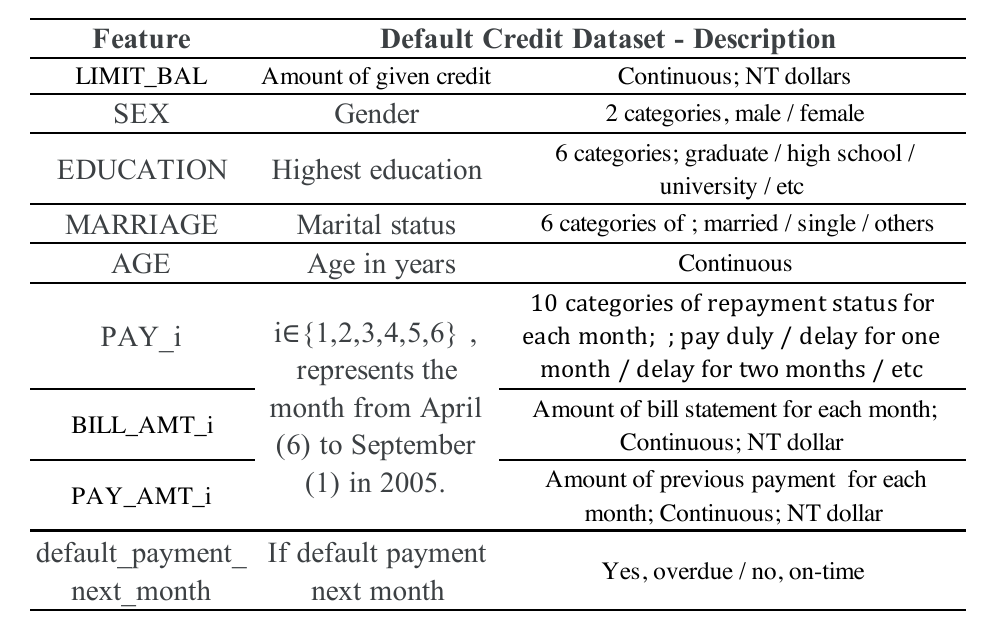}
\label{tab:app-default-dataset}
\end{center}
\end{table*}

\begin{table*}[t]
\begin{center}
\caption{Adult Income Dataset Description}
\includegraphics[width=0.95\textwidth]{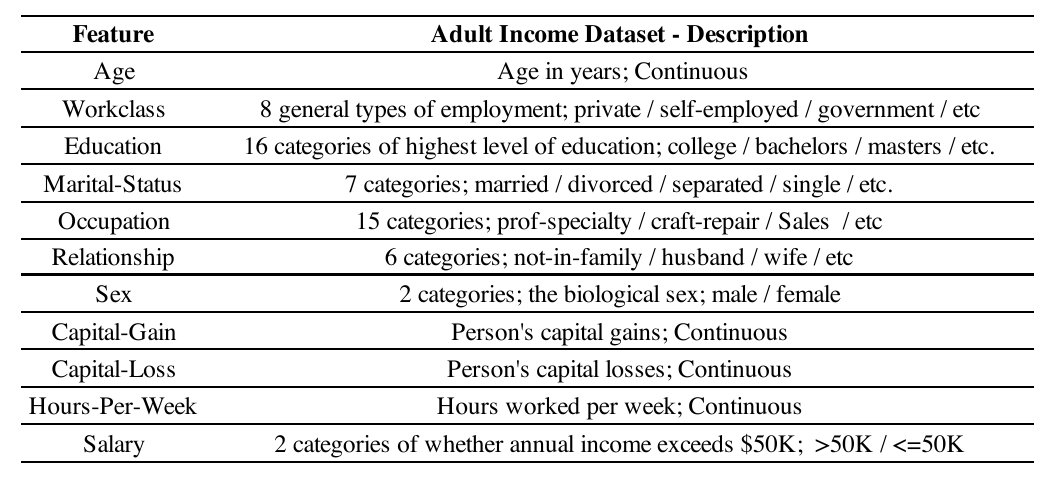}
\label{tab:app-adult-dataset}
\end{center}
\end{table*}

\section{Prompt Architecture}
\label{app:prompt}
We consider both zero-shot learning and few-shot learning (in-context learning).
Zero-shot strategy combines task description and question as its prompt content without providing examples. Few-shot strategy includes three roles, and the in-context content is generated based on selected K-demonstrations using different strategies (Table \ref{tab:sample-selection}). The default value of K is set to 8, the case of K=4 is disscussed in Section \ref{sec:fairer-mitigation}.
Table \ref{tab:demo_temp} and \ref{tab:demo_temp2} provide templates for few-shot learning in the Adult and Credit datasets respectively.

\begin{table*}[t]
\begin{center}
\caption{Few-shot Learning Templates for Adult Dataset}
\includegraphics[width=0.9\textwidth]{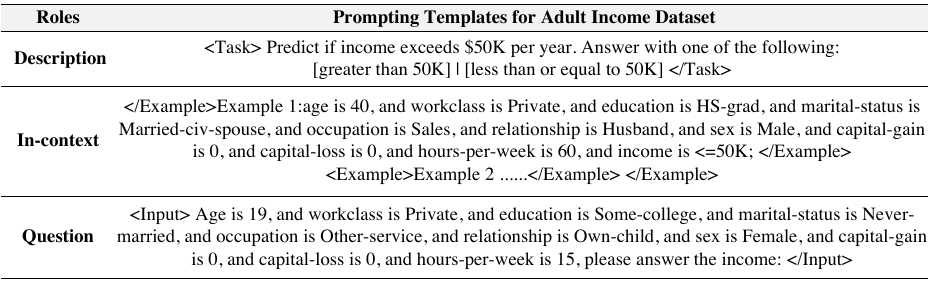}
\label{tab:demo_temp}
\end{center}
\end{table*}

\begin{table*}[t]
\begin{center}
\caption{Few-shot Learning Templates for Credit Dataset}
\includegraphics[width=0.9\textwidth]{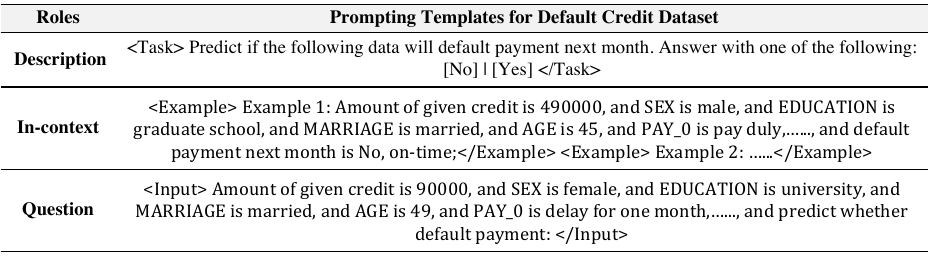}
\label{tab:demo_temp2}
\end{center}
\end{table*}

% \textbf{Few-shot Sample Selection Strategies
% }

% \usepackage{colortbl}

\begin{table*}[t]
\begin{center}
\caption{Demonstrations Selection Strategies}
\includegraphics[width=1\textwidth]{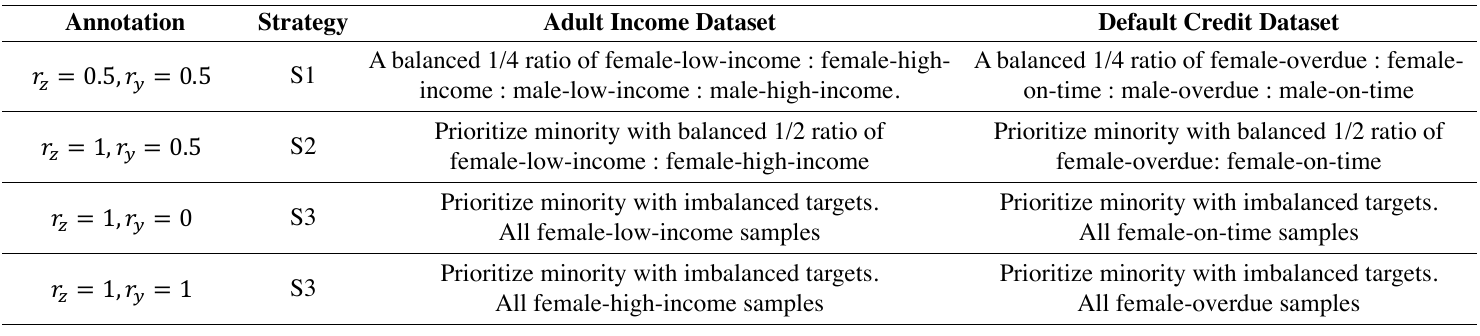}
\label{tab:sample-selection}
\end{center}
\end{table*}

\section{More Experimental Results}
\label{app:res}
Tables below present additional detailed results not listed in the main text.

\begin{table*}[t]
\begin{center}
\caption{Different LLMs performance on Default Credit Dataset}
\includegraphics[width=1\textwidth]{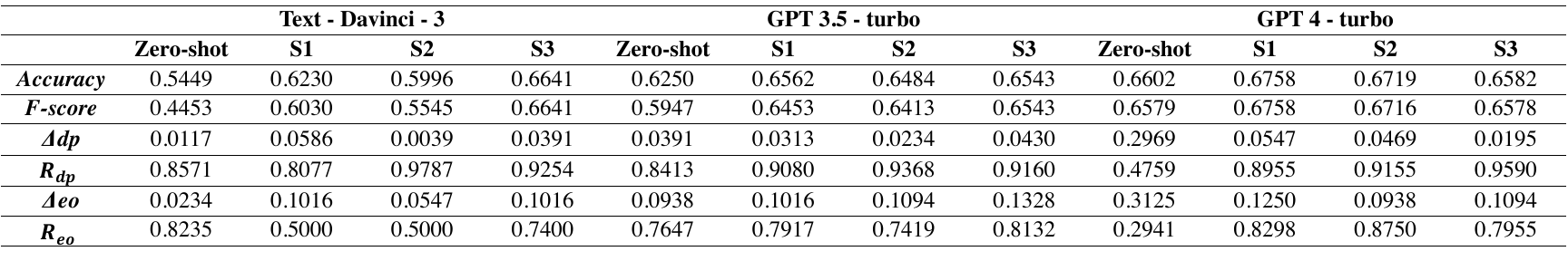}
\label{tab:app-diff-default}
\end{center}
\end{table*}

\begin{table*}[t]
\caption{Performance of Claude-3-haiku and Claude-3-sonnet with Zero-shot and Different Few-shot Strategies on Adult Dataset ($r_y=0.5$, K=4) }
\label{tab:my_label}
\centering
\begin{tabular}{ccccccccc}
\hline
& \multicolumn{4}{c}{Claude-3-haiku} & \multicolumn{4}{c}{Claude-3-sonnet} \\
\cline{2-9}
\rowcolor[HTML]{EFEFEF} \multicolumn{1}{c}{}   & Zero-shot & $r_z=0$     & $r_z=0.5$    & $r_z=1$     &  Zero-shot & $r_z=0$     & $r_z=0.5$    & $r_z=1$      \\
Accuracy                              & 0.7285    & 0.7070 & 0.7012 & 0.7031 & 0.6641 & 0.7266 & 0.7383 & 0.7520 \\
Precision                             & 0.7489    & 0.8118 & 0.7210 & 0.7047 & 0.6556 & 0.7302 & 0.7364 & 0.7699 \\
Recall                                & 0.6875    & 0.5391 & 0.6563 & 0.6992 & 0.6914 & 0.7188 & 0.7422 & 0.7188 \\
F-score                               & 0.7169    & 0.6479 & 0.6871 & 0.7020 & 0.6730 & 0.7244 & 0.7393 & 0.7434 \\
\rowcolor[HTML]{EFEFEF} \multicolumn{1}{c}{}   & Zero-shot & $r_z=0$     & $r_z=0.5$    & $r_z=1$     &  Zero-shot & $r_z=0$     & $r_z=0.5$    & $r_z=1$      \\
$\Delta dp$       & 0.2305    & 0.1797 & 0.1836 & 0.1563 & 0.0625 & 0.1484 & 0.1094 & 0.1445 \\
$R_{dp}$    & 0.5986    & 0.5741 & 0.6643 & 0.7279 & 0.8881 & 0.7379 & 0.8042 & 0.7319 \\
$\Delta eo$     & 0.2344    & 0.2188 & 0.2031 & 0.1641 & 0.0703 & 0.1563 & 0.1094 & 0.1719 \\
$R_{eo}$         & 0.3409    & 0.2800 & 0.5116 & 0.5625 & 0.8235 & 0.5455 & 0.6585 & \textbf{0.5714} \\
\hline
\small
\end{tabular}
\end{table*}

\begin{table*}[h]
\begin{center}
\caption{Performance of GPT3.5-turbo on the Adult Dataset through Few-shot Strategies (S1) with 5 random seeds}
\includegraphics[width=0.95\textwidth]{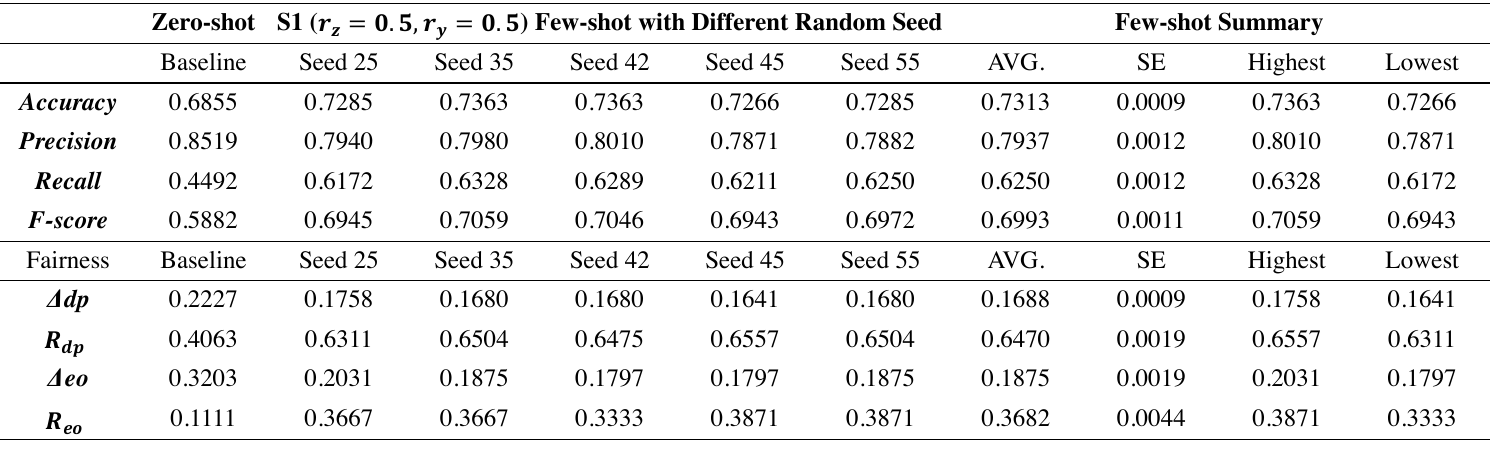}
\end{center}
% \label{tab:pertu_ch_pred_label}
\end{table*}

\begin{table*}[h]
\caption{Performance of GPT3.5-turbo on the Adult Dataset through Few-shot Strategies (S2) with 5 random seeds}
\begin{center}
\includegraphics[width=0.95\textwidth]{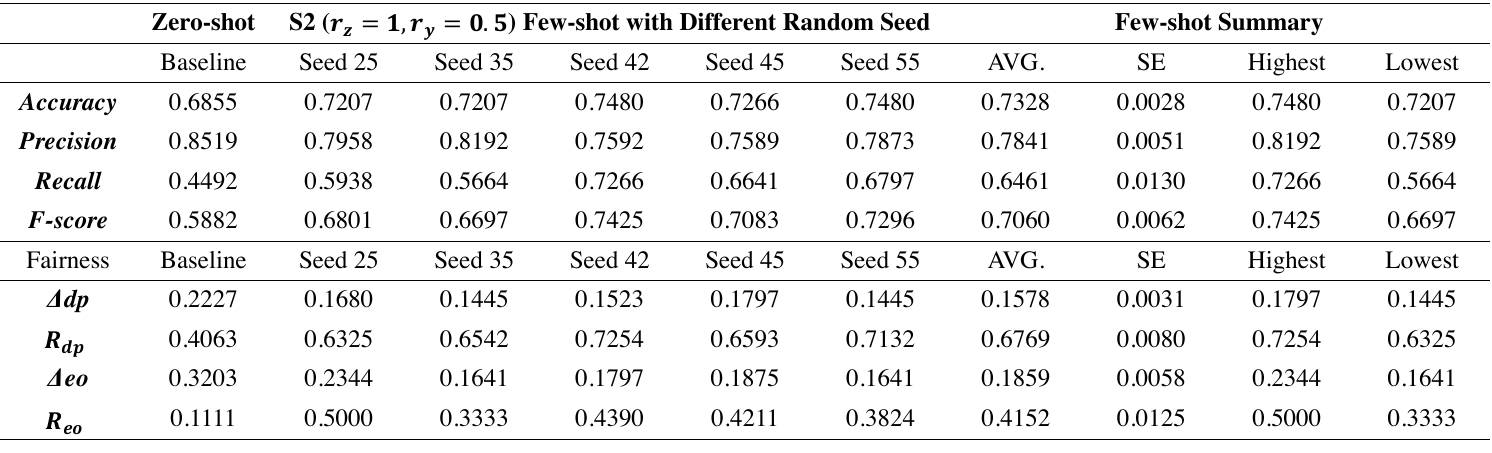}
\end{center}
% \label{tab:pertu_ch_pred_label}
\end{table*}

\begin{table*}[h]
\caption{Performance of GPT3.5-turbo on the Adult Dataset through Few-shot Strategies (S3) with 5 random seeds}
\begin{center}
\includegraphics[width=0.95\textwidth]{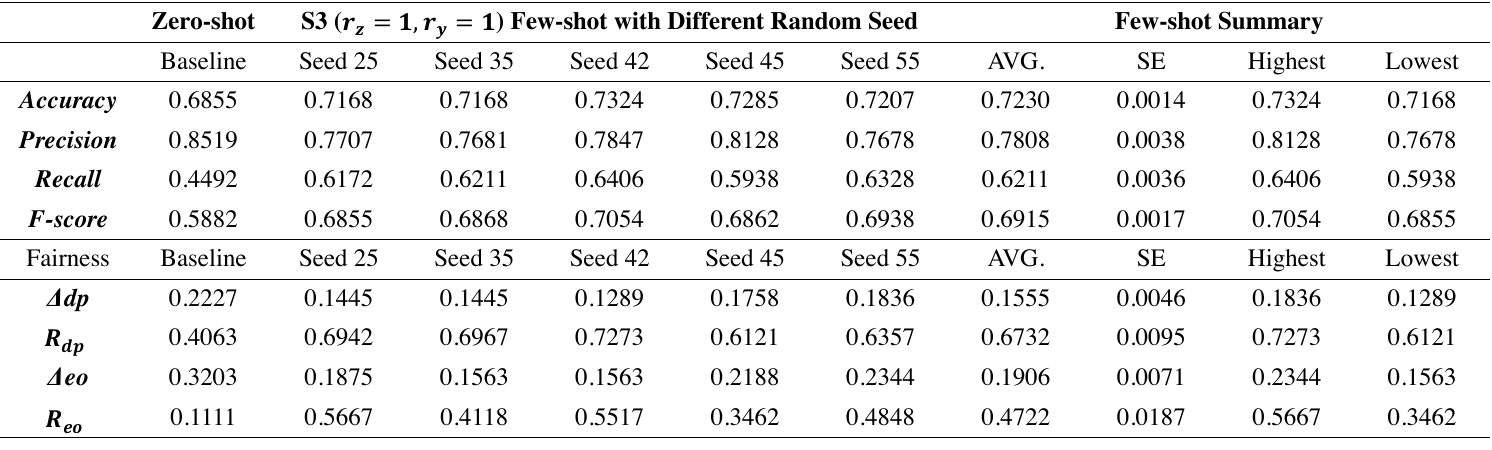}
\end{center}
% \label{tab:pertu_ch_pred_label}
\end{table*}

\begin{table*}
\caption{The Ablation Study with Balanced Labels in Minority Group (S2) under FCG Selections on GPT-3.5-turbo. The corresponding subgroups are denoted as $g_1(z=1,y=0)$ and $g_2(z=1,y=1)$.}
\begin{center}
\resizebox{0.8\textwidth}{!}{
\begin{tabular}{ccccccc}
\toprule
       K-shots       & \multicolumn{3}{c}{K=8}           & \multicolumn{3}{c}{K=4}           \\
\midrule
$g_2$  & Top K/2 & Top K/2    & Random K/2 & Top K/2 & Top K/2    & Random K/2 \\
\midrule
$g_1$  & Top K/2 & Random K/2 & Top K/2    & Top K/2 & Random K/2 & Top K/2    \\
\midrule
\rowcolor[HTML]{EFEFEF}
       & FCG ($g1+g2$)     & $FCG (g_2)$  & FCG ($g_1$)  & FCG ($g1+g2$)     & FCG ($g_2$)  & FCG ($g_1$) \\
\textbf{$Accuracy$ }  & 0.7793  & 0.7500     & 0.7559     & 0.7656  & 0.7441     & 0.7539     \\
\textbf{$Precision$}  & 0.7360  & 0.7013     & 0.7079     & 0.7194  & 0.7036     & 0.7031     \\
\textbf{$Recall$ }    & 0.8711  & 0.8711     & 0.8711     & 0.8711  & 0.8438     & 0.8789     \\
\textbf{$F score$ }   & 0.7979  & 0.7770     & 0.7811     & 0.7880  & 0.7673     & 0.7813     \\
\rowcolor[HTML]{EFEFEF}
       % & $FCG (g_1+g_2)$     & $FCG (g_2)$  & $FCG (g_1)$  & $FCG (g_1+g_2)$     & $FCG (g_2)$  & $FCG (g_1)$  \\
       & $FCG (g_1+g_2)$     & $FCG (g_2)$  & $FCG (g_1)$  & $FCG (g_1+g_2)$     & $FCG (g_2)$  & $FCG (g_1)$  \\
\textbf{$\Delta_{dp}$ }    & 0.0664  & 0.1172     & 0.1211     & 0.0859  & 0.1133     & 0.1016     \\
\textbf{$R_{dp}$}     & 0.8938  & 0.8276     & 0.8208     & 0.8675  & 0.8274     & 0.8497     \\
\textbf{$\Delta_{eo}$ }    & 0.1094  & 0.1328     & 0.1719     & 0.1172  & 0.1172     & 0.1328     \\
\textbf{$R_{eo}$ }    & 0.7021  & 0.6964     & 0.6140     & 0.7059  & 0.7170     & 0.6964    \\
\bottomrule
\end{tabular}
}
\end{center}
\end{table*}

\section{Discussion between Our FCG and LENs Algorithm}
\begin{figure*}[t]
\begin{center}
\includegraphics[width=0.9\textwidth]{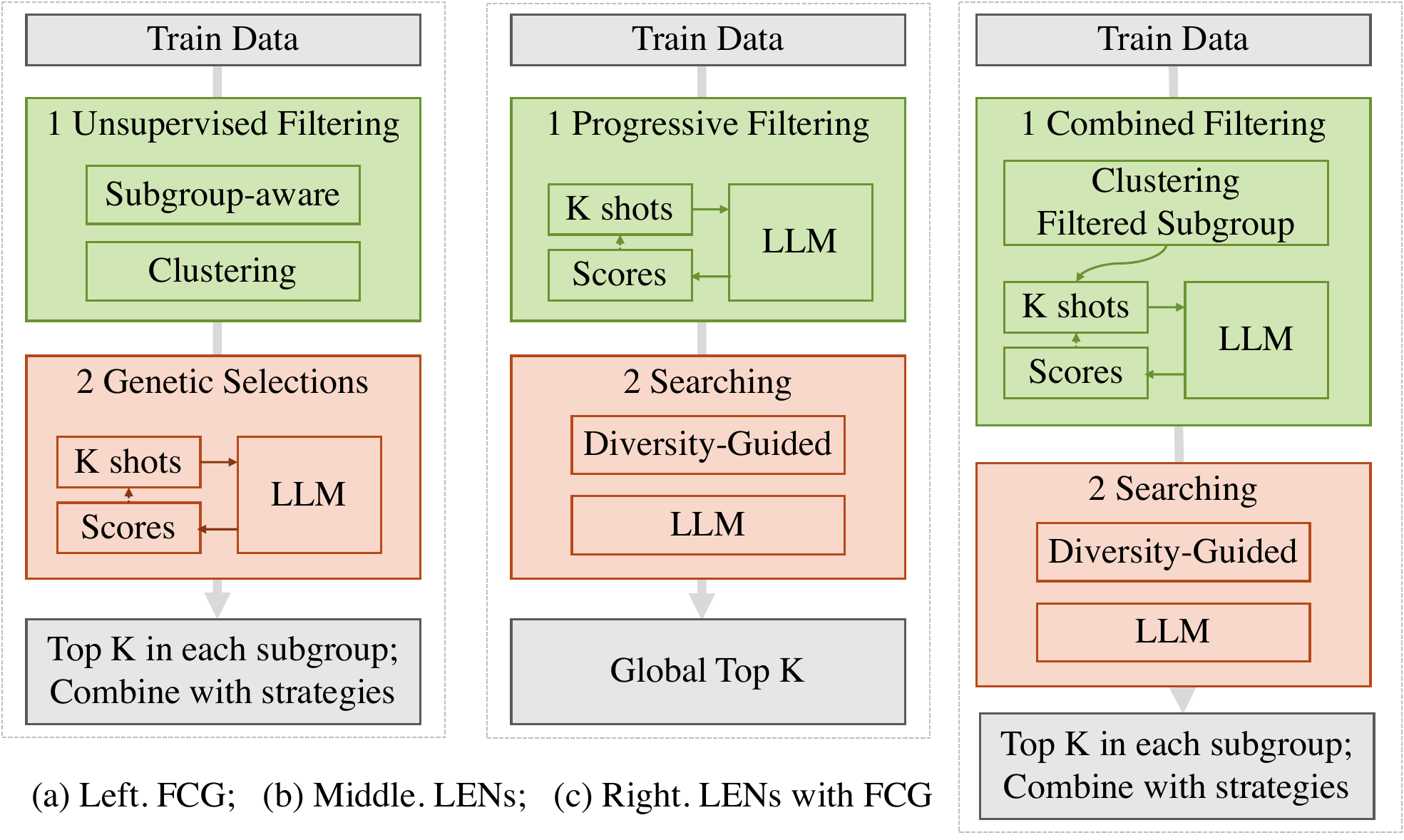}
\caption{The workflow comparison of demonstration selection algorithms: FCG (Ours, proposed in Section \ref{sec:fairer-mitigation}), LENs, and LENs with FCG.}
\label{fig:cmp_fcg_lens}
\end{center}
% \vspace{-10pt}
\end{figure*}

Our proposed FCG shares a similar architecture with LENs, another demonstration selection method. Here, we discuss these two methods and further explore the possibility of combining them. Given the time consumption of LENs, we simplified it by setting batch size to 8, and template index to \{0,1\}. The training dataset is randomly split ($seed=42$) into groups of 500 samples for parallel computation, with other settings kept at their defaults. For LENs with FCG, we follow our FCG parameters setting: first extract 160 representative samples by clustering, then perform LENs to find the final candidates.
Figure \ref{fig:cmp_fcg_lens} presents the overall workflow of FCG, LENs, and their possible combination. Both FCG and LENs involve two steps: (1) selecting partial data and (2) searching for the optimal based on the filtered data. There are two key differences in implementations.

\textbf{Supervised \& Unsupervised} LENs algorithm uses LLMs as the classification assistance in both stages. This is a straightforward and effective way. However, since the processing time of language models is related to the amount of information in the input text, the selection time can become very long when the input data is lengthy. 
This study focuses on tabular datasets, which have longer text when converted into prompts compared to commonly used NLP datasets. Therefore, we consider to optimize the method to reduce processing time and improve efficiency. Our FCG replaces LLMs with simpler unsupervised algorithms in the first stage. 
On the adult dataset, LENs can take over 50 hours to find supportive demonstrations ($batch size=8$), while FCG takes less than 3 hours ($K=4$). The result in Table \ref{tab:lens_fcg} validates the effectiveness: even if LLMs are not used initially, using LLMs to search on the validation set in the second stage can still find demonstrations that improve the model's prediction.

\textbf{Fairness Awareness} Another difference is that LENs use accuracy as the sole evaluation metric when selecting demonstrations. Our FCG takes sensitive features into account and selects demonstrations at the subgroup level. Additionally, FCG considers both accuracy and fairness metrics as constraints when calculating performance scores. Table \ref{tab:lens_fcg} confirms FCG with minority demonstrations prioritised strategy ($r_z=1$) shows fairer performance than baseline. 

Furthermore, we extend LENs with FCG (as shown on the right side of Figure \ref{fig:cmp_fcg_lens}) to make it fairness-aware. Table \ref{tab:lens_fcg} proves the effectiveness of this combination and also shows the best performance achieved when using more minority demonstrations ($r_z=1$).

\begin{table*}[t]
\caption{ICL Performance of LLaMa-3-8b on the Adult Dataset ($r_y=0.5$) using Different Demonstration Retrieval Methods (LENs, FCG, and Combined). }
\label{tab:lens_fcg}
\centering
\resizebox{0.76\textwidth}{!}{%
\begin{tabular}{cc|ccc|ccc}
\hline
LLaMa-3 & LENs (K = 4) & \multicolumn{3}{c|}{FCG (Ours, K = 4)} & \multicolumn{3}{c}{LENs with FCG (K = 4)} \\ 
%\hline
\rowcolor[HTML]{EFEFEF}  & Baseline & \textbf{$r_z=0$} & \textbf{$r_z=0.5$} & \textbf{$r_z=1$} & \textbf{$r_z=0$} & \textbf{$r_z=0.5$} & \textbf{$r_z=1$} \\ 
\textbf{$Accuracy$} & 0.6270 & 0.6406 & 0.6680 & \textbf{0.7148} & 0.5957 & \textbf{0.6543} & 0.6504 \\ %\cline{2-8}
\textbf{$Precision$} & 0.7211 & \textbf{0.8462} & 0.7389 & 0.7778 & 0.7426 & \textbf{0.7687} & 0.6103 \\ 
%\cline{2-8}
\textbf{$Recall$} & 0.4141 & 0.3438 & 0.5195 & \textbf{0.6016} & 0.2930 & 0.4414 & \textbf{0.8320} \\ %\cline{2-8}
\textbf{$F-score$} & 0.5261 & 0.4889 & 0.6101 & \textbf{0.6784} & 0.4202 & 0.5608 & \textbf{0.7041} \\
%\cline{2-8}
\rowcolor[HTML]{EFEFEF}  & Baseline & \textbf{$r_z=0$} & \textbf{$r_z=0.5$} & \textbf{$r_z=1$} & \textbf{$r_z=0$} & \textbf{$r_z=0.5$} & \textbf{$r_z=1$} \\ 
\textbf{$\Delta dp$} & 0.1523 & 0.1250 & 0.1094 & \textbf{0.0938} & 0.1602 & 0.1445 & \textbf{0.0039} \\ %\cline{2-8}
\textbf{$R_{dp}$} & 0.5806 & 0.5294 & 0.7308 & \textbf{0.7838} & 0.4225 & 0.5978 & \textbf{0.9943} \\ %\cline{2-8}
\textbf{$\Delta eo$} & 0.2344 & 0.1719 & 0.1172 & \textbf{0.0938} & 0.2109 & 0.1953 & \textbf{0.0469} \\ %\cline{2-8}
\textbf{$R_{eo}$} & 0.5588 & 0.2308 & 0.5161 & \textbf{0.5714} & 0.3000 & 0.4783 & \textbf{0.9155} \\ \hline
\end{tabular}%
}
\end{table*}

\end{document}